\pdfoutput=1
\documentclass[11pt]{article}

\usepackage[final]{acl}
\usepackage{times}
\usepackage{comment,multirow}
\usepackage{makecell}
\usepackage{amssymb}
\usepackage[most]{tcolorbox}
\usepackage{caption}
\usepackage{xspace}

\definecolor{myblue}{rgb}{0.631, 0.682, 0.765}
\definecolor{myorange}{rgb}{0.749, 0.376, 0.0}
\newcommand{\DeAR}{\textsc{DeAR}\xspace}

 \usepackage{booktabs} 
\usepackage{graphicx}
\usepackage{listings}
\usepackage{subcaption}
\newcommand{\code}[1]{\texttt{#1}}
\title{ DeAR: Dual-Stage Document Reranking with Reasoning Agents via LLM Distillation }

\author{
    \textbf{Abdelrahman Abdallah, Jamshid Mozafari,   Bhawna Piryani,  Adam Jatowt} \\
    University of Innsbruck \\
    \texttt{\{abdelrahman.abdallah, jamshid.mozafari, bhawna.piryani,} \\
    \texttt{  adam.jatowt\}@uibk.ac.at}
}

\begin{document}
\maketitle
\begin{abstract}

Large Language Models (LLMs) have transformed listwise document reranking by enabling global reasoning over candidate sets, yet single models often struggle to balance fine-grained relevance scoring with holistic cross-document analysis. We propose \textbf{De}ep\textbf{A}gent\textbf{R}ank (\textbf{\DeAR}), an open-source framework that decouples these tasks through a dual-stage approach, achieving superior accuracy and interpretability. In \emph{Stage 1}, we distill token-level relevance signals from a frozen 13B LLaMA teacher into a compact \{3, 8\}B student model using a hybrid of cross-entropy, RankNet, and KL divergence losses, ensuring robust pointwise scoring. In \emph{Stage 2}, we attach a second LoRA adapter and fine-tune on 20K GPT-4o-generated chain-of-thought permutations, enabling listwise reasoning with natural-language justifications. Evaluated on TREC-DL19/20, eight BEIR datasets, and NovelEval-2306, \DeAR surpasses open-source baselines by +5.1 nDCG@5 on DL20 and achieves 90.97 nDCG@10 on NovelEval, outperforming GPT-4 by +3.09. Without fine-tuning on Wikipedia, DeAR also excels in open-domain QA, achieving 54.29 Top-1 accuracy on Natural Questions, surpassing baselines like MonoT5, UPR, and RankGPT. 
Ablations confirm that dual-loss distillation ensures stable calibration, making \DeAR a highly effective and interpretable solution for modern reranking systems.\footnote{Dataset and code available at \url{https://github.com/DataScienceUIBK/DeAR-Reranking}.}.
\end{abstract}

\section{Introduction}
\begin{figure}[htbp]
    \centering
    \includegraphics[width=0.5\textwidth]{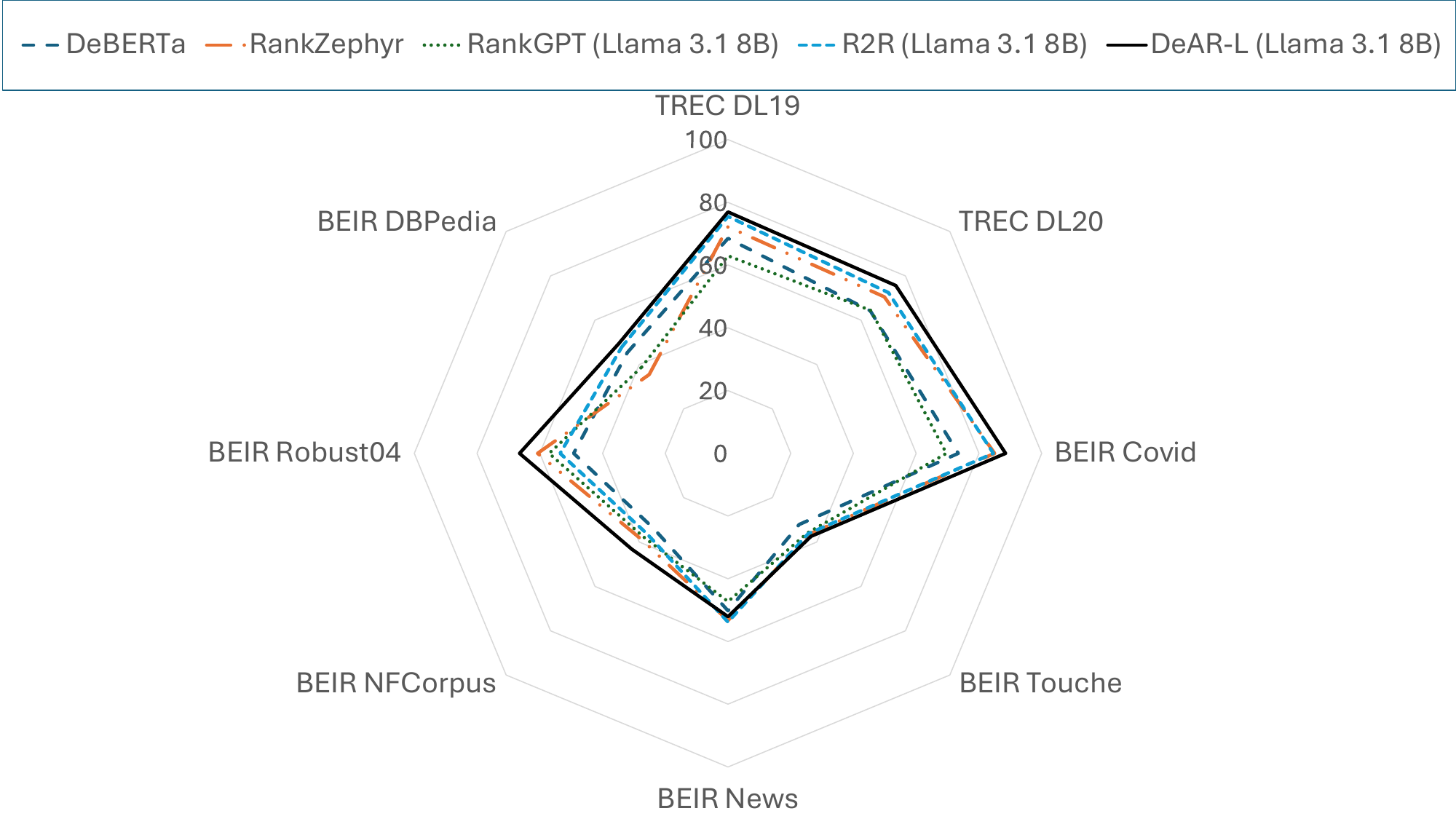}
    \caption{Radar chart comparing nDCG@5 performance of top reranking methods, including DeBERTa, RankZephyr, RankGPT (LLaMA 3.1 8B), R2R (LLaMA 3.1 8B), and \DeAR-L (LLaMA 3.1 8B), across TREC DL19 and BEIR datasets (Covid, NFCorpus, Touche, DBPedia, News, Robust04).}
    \label{fig:radar_chart_all_beir}
\end{figure}

\begin{figure*}[t]
\centering

%\begin{figure*}
%\begin{subfigure}[t]{0.5\textwidth}
\begin{subfigure}[t]{1\textwidth}
\footnotesize
\centering
\resizebox{.9\textwidth}{!}{%
\begin{tcolorbox}[colback=blue!0!white, colframe=orange!75!black, title=Synthetic training sample,fonttitle=\bfseries]
\textbf{System Prompt:} You are RankLLM, an intelligent assistant that can rank passages based on their relevancy to the query.

\vspace{0.5em}
\textbf{Instruction:} I will provide you with 3 passages, each indicated by a numerical identifier []. Rank the passages based on their relevance to the search query: \textit{what is a shape of art}.

{[1]} In the visual arts, shape is a flat, enclosed area of an artwork created through line, texture, colour or an area enclosed by other shapes. {[...]}\\
{[2]} A shape in art is a closed line that is limited to two directions: width and length. {[...]}\\
{[3]} Form is a three-dimensional geometrical figure (i.e., sphere, cube, cylinder, etc.), as opposed to a shape, which is two-dimensional. {[...]}

\noindent \textbf{Search Query:} what is a shape of art

\noindent \textbf{Steps to follow:}
\begin{enumerate}
  \item List the information requirements to answer the query.
  \item For each requirement, find the passages that include the relevant information.
  \item Rank the passages in descending order of relevance using only the identifiers (e.g., [2] $>$ [1]).
\end{enumerate}
The format of the final output should be ‘ \#\#\# Final Reranking: {[]} $>$ {[]}’, " "e.g., \#\#\# Final Reranking: {[2]} $>$ {[1]}. \\
\vspace{0.5em}

% \textcolor{orange}{\texttt{<think>}}\\
% \textbf{Step 1}: The query \textit{“what is a shape of art”} requires:
% \begin{itemize}
%   \item Definition of shape in art.
%   \item Characteristics of shape (e.g., geometric vs. organic).
%   \item Distinction between shape and form.
% \end{itemize}

% \textbf{Step 2}: Identify supporting passages:
% \begin{itemize}
%   \item Requirement 1: [1], [2] both define shape; [3] does not.
%   \item Requirement 2: [2] discusses characteristics; [1] minimally touches this; [3] does not.
%   \item Requirement 3: [1], [3] both distinguish shape vs. form; [2] does not.
% \end{itemize}

% \textbf{Step 3}: Ranking:
% \begin{itemize}
%   \item [1] covers definition and shape vs. form.
%   \item [2] covers definition and characteristics.
%   \item [3] only briefly contrasts form and shape.
% \end{itemize}

% Final Ranking: [1] > [2] > [3]\\
% \textcolor{orange}{\texttt{</think>}}\\
\noindent \textbf{\#\#\# Final Reranking: } [1] $>$ [2] $>$ [3]
\end{tcolorbox} }
%\caption{Training example from our sampled synthetic RankLLM chains. Passages and reasoning steps are abbreviated for brevity.}

\end{subfigure}

\caption{Illustration of RankLLM training components. Synthetic training example with reasoning and final reranking. Passages and reasoning steps are abbreviated for brevity.}
\label{fig:rankllm-train-example}
\end{figure*}

%%%%%%%%%%%%%%%%%%%%%%%

Document reranking refines top candidate documents retrieved by a first-stage system to improve relevance to a user query. It plays a critical role in tasks like web search~\cite{bajaj2016ms,abdallah2025tempretriever}, open-domain QA~\cite{chen2017reading,gruber2024complextempqa}, fact verification~\cite{thorne2018fever}, and Retrieval-Augmented Generation (RAG)~\cite{lewis2020retrieval}, where ranking quality directly impacts downstream results. Transformer-based models (e.g., BERT~\cite{devlin2019bert}, T5~\cite{raffel2020exploring}) and instruction-tuned LLMs (e.g., InstructGPT~\cite{ouyang2022training}, LLaMA~\cite{touvron2023llama}, GPT-4~\cite{achiam2023gpt}) have driven reranking progress. While pointwise~\cite{sachan2022improving,abdallah2025asrank,abdallah2025dynrank} and pairwise~\cite{qin2023large} approaches dominate earlier work, listwise reranking with LLMs~\cite{sun2023chatgpt,pradeep2023rankvicuna} offers global ranking benefits. However, these often rely on expensive proprietary APIs and suffer from context-length limitations and brittle reasoning. As shown in Figure~\ref{fig:radar_chart_all_beir}, our proposed \DeAR-L achieves superior nDCG@5 performance across diverse datasets, outperforming baselines like RankZephyr and RankGPT.

To address these challenges, open-source efforts have explored knowledge distillation (KD) to transfer ranking abilities from large LLMs to smaller, more efficient student models~\cite{sun2023chatgpt,pradeep2023rankvicuna}. However, existing methods commonly depend on synthetic listwise permutations that risk propagating teacher errors such as hallucinations or misrankings. This motivates two key research questions: (1) \emph{Can we balance KL divergence with ranking loss to effectively distill logit-level signals from the teacher while mitigating noise?} (2) \emph{Can we incorporate synthetic reasoning chains to retain listwise reasoning benefits without overloading model context windows?}

We address these questions with a central insight: reranking performance can be enhanced by guiding the student with reasoning chains of thought (CoT) derived from synthetic data. We introduce \textbf{\DeAR}, a novel dual-stage reranking framework that combines pointwise and listwise learning with reasoning-augmented supervision. Built on a frozen LLM backbone (e.g., LLaMA-13B) with lightweight LoRA adapters~\cite{hu2022lora}, \DeAR trains on 20,000 synthetic reasoning examples and achieves performance comparable to GPT-4o, while surpassing RankZephyr in inference efficiency (see Section~\ref{sec:time-ndgc}).

%In \emph{Stage 1}, we perform pointwise reranking using logit-level KD, where the student learns calibrated document scores by combining cross-entropy, RankNet, and KL divergence losses. This avoids reliance on synthetic permutations. In \emph{Stage 2}, we add a second LoRA adapter and fine-tune the model on listwise supervision, guided by GPT-4o-generated reasoning chains that explicitly decompose the ranking process—identifying query requirements, aligning passages, and ordering results. This enables holistic reasoning over smaller document sets while avoiding long-context inefficiencies.

%Our dual-stage design provides several benefits. Stage 1 preserves the teacher’s confidence distribution without noisy labels. Stage 2 improves interpretability by teaching the model \emph{what} to rank and \emph{why}, aligning with recent trends in explainable LLM reasoning~\cite{Magister2022TeachingSL,guo2025deepseek}.

%\vspace{1ex}
\noindent\textbf{Our contributions are as follows:}
\begin{itemize}
\item We propose \textbf{\DeAR}, a dual-stage reranking framework that integrates pointwise cross-entropy learning with reasoning-augmented listwise ranking.%, achieving high accuracy and interpretability.
\item We introduce a teacher–student pipeline where a LLM-based teacher transfers relevance signals to a student via logit-level distillation, combining cross-entropy, RankNet, and KL divergence losses.
\item We construct 20$K$ synthetic ranking examples with CoT reasoning.%, enabling listwise training with interpretable supervision.
\item Extensive experiments on DL19, DL20, and BEIR-6 show that \DeAR matches or outperforms larger baselines (e.g., GPT-4o, RankZephyr), improving NDCG@5 by up to +5.1 on DL20 while remaining lightweight and fast.
\end{itemize}

\begin{figure*}[h]
\centering
\includegraphics[width=\textwidth]{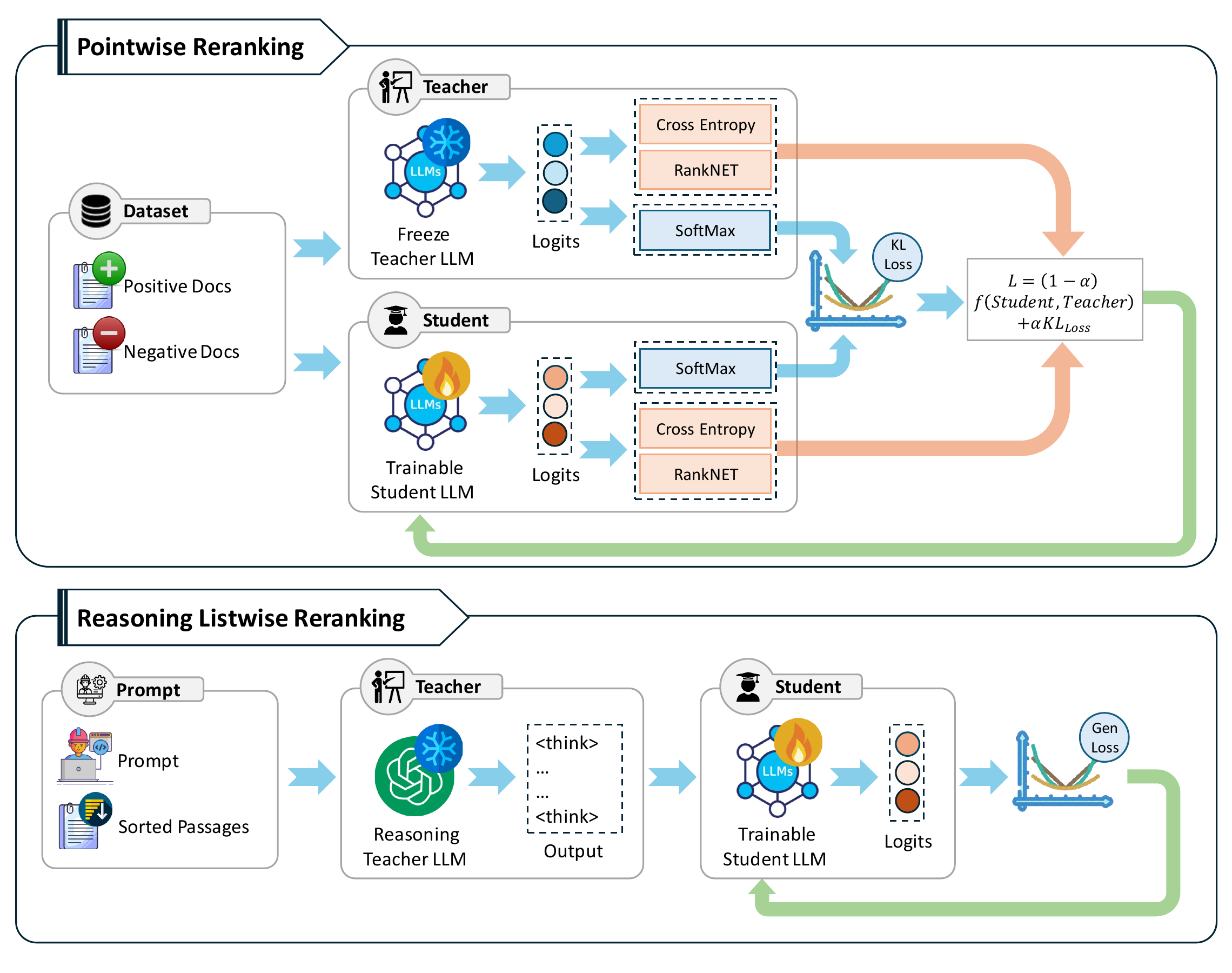}

\caption{Overview of the \textbf{\DeAR} dual-stage training pipeline. 
\textbf{Top:} In the pointwise stage, a frozen teacher LLM generates relevance logits for positive/negative documents, which are distilled into a student model using cross-entropy, RankNet, and KL divergence losses.
\textbf{Bottom:} In the listwise stage, a reasoning teacher produces step-by-step chain-of-thought explanations and ranked outputs over candidate sets. The student is trained to generate coherent reasoning and rankings via generation loss.}
    \label{fig:training}
\end{figure*}
\section{Related Work}

Reranking methods are typically pointwise, pairwise, or listwise. Pointwise models like monoBERT~\cite{nogueira2020document} score each document independently using pretrained transformers~\cite{devlin2019bert}, but ignore inter-document context~\cite{sachan2022improving}. Pairwise approaches (e.g., duoBERT, duoT5~\cite{pradeep2021expando}) compare document pairs to infer preferences, at the cost of efficiency~\cite{qin2023large}. Listwise methods, powered by LLMs like GPT-3.5 and GPT-4, enable zero-shot ranking through prompting~\cite{sun2023chatgpt}, though reliance on proprietary APIs limits reproducibility. Open-source variants like RankVicuna and RankZephyr~\cite{pradeep2023rankzephyr} address this via distilled listwise models, but remain sensitive to hallucinations and prompt ordering. 

Knowledge distillation (KD) compresses large models into smaller ones by transferring logits or intermediate signals~\cite{hinton2015distilling}. In IR, it has been used for both retrieval~\cite{guo2021distilling} and reranking~\cite{wang2021knowledge}. RankGPT and RankZephyr apply permutation-based KD using teacher rankings, but such discrete labels may discard confidence information. 

Other work explores reasoning distillation~\cite{Magister2022TeachingSL,Fu2023SpecializingSL}, where small models benefit from teacher-generated explanations. Our method extends this by combining logit-level KD (Stage 1) with reasoning-based listwise training (Stage 2), preserving fine-grained supervision while improving stability. Reasoning improves reranking accuracy and interpretability. R2R~\cite{ji2024reasoningrank} distills direct and comparative explanations from GPT-4 for MSMARCO and BEIR. RankGPT also leverages implicit LLM reasoning via prompt completions. 

\section{Method}

\subsection{Preliminaries}
\textbf{Task Definition.} Given a query $q \in \mathcal{Q}$ and a corpus $\mathcal{C} = \{d_1, \ldots, d_n\}$, the reranking task aims to reorder a top-$k$ candidate set ($k \ll n$), initially retrieved by a first-stage bi-encoder~\citep{karpukhin2020dense}, to maximize relevance to $q$. The reranker refines these candidates using a more expressive model to optimize metrics like nDCG~\citep{wang2013theoretical,jarvelin2002cumulated}.%applied to tasks such as web search~\citep{bajaj2016ms}, QA~\citep{chen2017reading}, and RAG~\citep{lewis2020retrieval}.

\textbf{Pointwise vs. Listwise Reranking.} Pointwise rerankers score each $(q, d)$ pair independently via $f(q, d)$~\citep{sachan2022improving}, offering efficiency but no inter-document reasoning. We improve this via knowledge distillation from LLMs to LoRA-equipped student models~\citep{hu2022lora}. Listwise methods (e.g., ListNet~\citep{cao2007learning}, LambdaLoss~\citep{burges2010ranknet}) model cross-document interactions~\citep{ma2023zero,pradeep2023rankzephyr}, but suffer from input order sensitivity and transformer context limits~\citep{sun2023chatgpt}.

\subsection{Pointwise Reranking with KL Distillation}

The first stage of \DeAR employs a pointwise reranking approach, leveraging KD to transfer the teacher model’s nuanced relevance judgments to a compact student model. This stage produces calibrated relevance scores for individual documents, forming the foundation for the subsequent listwise refinement.  For a query $q_i$ and document $d_{ij}$ from the candidate set $D_i = \{d_{i1}, \ldots, d_{im}\}$, the input is formatted as:
\begin{equation}
s_{ij} = [\text{query: } q_i, \text{ document: } d_{ij}, \texttt{</s>}],
\end{equation}
where $\texttt{</s>}$ denotes the end-of-sequence token. Both the student model $f_s(\cdot; \theta_s)$ and teacher model $f_t(\cdot; \theta_t)$, implemented as transformer-based sequence classifiers, process $s_{ij}$ through $L$ layers of self-attention and feed-forward networks. Let $\mathbf{H}^l \in \mathbb{R}^{l \times d}$ represent the hidden states at layer $l$ for sequence length $l$ and hidden dimension $d$. The final hidden state corresponding to $\texttt{</s>}$, denoted $\mathbf{h}^L_{\texttt{</s>}} \in \mathbb{R}^d$, is projected to a scalar relevance score:
\begin{align}
\hat{y}_{ij}^s &= \mathbf{w}_s^\top \mathbf{h}^L_{\texttt{</s>}}(s_{ij}; \theta_s) + b_s, \\
\hat{y}_{ij}^t &= \mathbf{w}_t^\top \mathbf{h}^L_{\texttt{</s>}}(s_{ij}; \theta_t) + b_t,
\end{align}
where $\mathbf{w}_s, \mathbf{w}_t \in \mathbb{R}^d$ and $b_s, b_t \in \mathbb{R}$ are learnable projection parameters. 
For a batch of $B$ queries, each associated with $m$ documents, the student and teacher produce score matrices $\mathbf{S}^s, \mathbf{S}^t \in \mathbb{R}^{B \times m}$, where $\mathbf{S}^s_{bj} = \hat{y}_{ij}^s$ and $\mathbf{S}^t_{bj} = \hat{y}_{ij}^t$ for query $q_i$ (batch index $b$) and document $d_{ij}$ (index $j$). The training objective combines a ranking loss with a KD loss, parameterized by a weighting factor $\alpha \in [0, 1]$.

We support multiple ranking loss functions to accommodate different supervision signals: Softmax Cross-Entropy (PointCE): Treating reranking as a classification task, we assign binary labels $\mathbf{y}_i = (y_{i1}, \ldots, y_{im})$, where $y_{ij} = 1$ for the most relevant document and $y_{ij} = 0$ otherwise. The loss is:
\begin{align}
\mathcal{L}_{\text{PointCE}}(\mathbf{y}_i, \mathbf{S}^s_i) &= -\sum_{j | y_{ij} = 1} \log(\sigma(\mathbf{S}^s_{ij})) \notag \\
&\quad - \sum_{j | y_{ij} = 0} \log(1 - \sigma(\mathbf{S}^s_{ij}))
\end{align}
where $\sigma(x) = (1 + e^{-x})^{-1}$ is the logistic function~\citep{nogueira2020document}.

RankNet Loss: To model pairwise preferences, we use the RankNet loss~\citep{burges2005learning}. Given relevance ranks $r_{ij}$ derived from $\mathbf{y}_i$ (lower $r_{ij}$ indicates higher relevance), the loss is:
\begin{align}
\mathcal{L}_{\text{RankNet}}(\mathbf{y}_i, \mathbf{S}^s_i) &= \sum_{j=1}^m \sum_{j'=1}^m \mathbb{1}_{r_{ij} < r_{ij'}} \notag \\
&\quad \cdot \log(1 + \exp(\mathbf{S}^s_{ij'} - \mathbf{S}^s_{ij}))
\end{align}

Knowledge Distillation Loss (KD) aligns the student’s softened score distribution with the teacher’s using KL divergence. For query $q_i$, scores $\mathbf{S}^s_i, \mathbf{S}^t_i \in \mathbb{R}^m$ are normalized with temperature $\tau$:
\begin{equation}
\mathbf{P}^s_i = \text{softmax}(\mathbf{S}^s_i / \tau), \quad \mathbf{P}^t_i = \text{softmax}(\mathbf{S}^t_i / \tau)
\end{equation}
The KD loss is:
\begin{align}
\mathcal{L}_{\text{KD}}(\mathbf{S}^s_i, \mathbf{S}^t_i) &= \tau^2 \cdot \text{KL}(\mathbf{P}^s_i || \mathbf{P}^t_i) \notag \\
&= \tau^2 \sum_{j=1}^m \mathbf{P}^s_{ij} \log \left( \frac{\mathbf{P}^s_{ij}}{\mathbf{P}^t_{ij}} \right)
\end{align}
The total loss combines the ranking loss and KD loss:
\begin{align}
\mathcal{L}_{\text{total}}(\mathbf{y}_i, \mathbf{S}^s_i, \mathbf{S}^t_i) &= (1 - \alpha) \cdot \mathcal{L}_{\text{rank}}(\mathbf{y}_i, \mathbf{S}^s_i) \notag \\
&\quad + \alpha \cdot \mathcal{L}_{\text{KD}}(\mathbf{S}^s_i, \mathbf{S}^t_i)
\end{align}
where $\mathcal{L}_{\text{rank}}$ is either $\mathcal{L}_{\text{PointCE}}$ or $\mathcal{L}_{\text{RankNet}}$, and $\alpha \in [0, 1]$ balances the losses. Figure~\ref{fig:training} illustrates the pointwise reranking stage of \DeAR, where the student model learns from the frozen teacher using a combination of cross-entropy, RankNet, and KL-divergence losses.  We train two student models: one with $\mathcal{L}_{\text{PointCE}} + \mathcal{L}_{\text{KD}}$ and another with $\mathcal{L}_{\text{RankNet}} + \mathcal{L}_{\text{KD}}$. The teacher provides logits without updates. LoRA ensures efficiency, producing a top-100 ranked list for the listwise stage.

\begin{table*}[!t]
\centering
\scriptsize
\setlength\tabcolsep{2pt}
\resizebox{\textwidth}{!}{
\begin{tabular}{l| cc | cc | cccccccc | c }

\toprule
\textbf{Method}& prev. & Top-$K$& DL19 & DL20 & Covid &  NFCorpus &  Touche &  DBPedia & SciFact &  Signal & News &  Robust04 & BEIR (Avg) \\

\midrule

BM25
& - & -& 50.58 & 47.96 & 59.47 & 30.75 & \textbf{44.22} & 31.80 & 67.89 & 33.05 & 39.52 & 40.70 & 43.42
\\

\midrule

\multicolumn{14}{c}{\textbf{Supervised}}\\ 

\midrule

monoBERT (340M)
& BM25 & 100 & 70.50 & 67.28 & 70.01 & 36.88 & 31.75 & 41.87 & 71.36 & 31.44 & 44.62 & 49.35 & 47.16
\\

monoT5 (220M)
& BM25 & 100 & 71.48 & 66.99 & 78.34 & 37.38 & 30.82 & 42.42 & 73.40 & 31.67 & 46.83 & 51.72 & 49.07
\\

monoT5 (3B)
& BM25 & 100 & 71.83 & 68.89 & 80.71 & 38.97 & 32.41 & 44.45 &  76.57 & 32.55 & 48.49 & 56.71 & 51.36
\\

Cohere Rerank-v2
& BM25 & 100 & 73.22 & 67.08 & 81.81 & 36.36 & 32.51 & 42.51 & 74.44 & 29.60 & 47.59 & 50.78 & 49.45
\\

%TART-Rerank (3B)
% &67.43 & 59.19 & 72.80 & 33.40 & 24.90 & 46.80 & 77.70 & - &- & -&-
% \\
%RankT5 (3B)
%& -& -& 80.71 &38.10 &44.01 & 44.22&  74.99&  32.00& - & -&
% \\

\midrule
\multicolumn{14}{c}{\textbf{Unsupervised}}\\ 
\midrule

UPR (3B)
& BM25 & 100& 53.85 & 56.02 & 68.11 & 35.04 & 19.69 & 30.91 & 72.69 & 31.91 & 43.11 & 42.43 & 42.99
\\

% SGPT-CE (6.1B)
% & 79.1 & 34.7 & 23.4 & 37.0 & 68.2 & 32.3 & 46.6 & 48.0 & 46.16
% \\

InPars (3B)
& - & 100 & - & 66.12 &  78.35 & - & - & - & - & - & - & - & -
\\

% HyDE (\code{text-davinci-003})
% & 59.3 & - & 27.3 & - & 44.0 & - & 46.6 & - & 36.8 & 69.1
% \\

% Promptagator++ (zero-shot)$^\dagger$
% & - & - & 76.0 & 36.0 & 27.8 & 41.3 & 73.6 & - & - & - & -
% \\

Promptagator++
& - & 100 & - & - & 76.2 & 37.0 & 38.1 & 43.4 & 73.1 & - & - & - & -
\\

%\midrule
%\multicolumn{5}{l}{LLM API (Permutation generation)}\\
%\midrule

RankGPT (llama 3.1 8B) & BM25 & 100 &58.46 & 59.68 & 69.61 & 33.62 & 37.98  & 37.25 & 69.82 & 32.95 &43.90  \\

RankGPT-3.5
& BM25 & 100 & 65.80 & 62.91  & 76.67 & 35.62 & 36.18 & 44.47 & 70.43 & 32.12 & 48.85 & 50.62 & 49.37
\\

RankGPT-4
& RankGPT-3.5 & 30 & 75.59 & 70.56 & 85.51 & 38.47 & 38.57 & 47.12 & 74.95 & 34.40 & 52.89 & 57.55 & 53.68
\\

% \code{gpt-4}$^\dagger$
% & \textbf{75.59} & \textbf{70.56} & \textbf{85.51} & \textbf{38.47} & \textbf{38.57} & \textbf{47.12} & \textbf{74.95} & \textbf{34.40} & \textbf{52.89} & \textbf{57.55} & \textbf{53.68}
% \\

\midrule
\multicolumn{14}{c}{ \textbf{\DeAR-Pointwise (\DeAR-P)} } \\
\midrule
 Llama3.1-8B (RL)$\dagger$ & BM25 & 100& 72.17 & 68.93 & 85.21 & 37.01 & 34.88 & 45.56 & 77.43 & 30.16 & 52.05 & 54.42 & 52.09 \\
 Llama3.1-8B (BC)$\ddagger$ & BM25 & 100& 74.50 & 68.71 & 84.14 & 36.57 & 37.23 & 46.27 & 77.39 & 29.91 & 51.71 & 52.43 & 51.95 \\
Llama3.1-3B (RL)$\S$ & BM25 & 100 & 72.94 & 69.21 & 83.01 & 36.30 & 35.76 & 45.96 & 74.45 & 28.64 & 50.84 & 49.78 & 50.59 \\
Llama3.1-3B (BC)$\P$ & BM25 & 100 & 74.49 & 69.02 & 82.91 & 35.78 & 36.17 & 45.28 & 75.48 & 29.14 & 48.99 & 50.93 & 50.58 \\

\midrule
\multicolumn{14}{c}{ \textbf{\DeAR-Listwise (\DeAR-L)} }\\
\midrule

GPT-4 &$\dagger$ & 30&75.74 &\textbf{ 72.18} & 86.28 & \textbf{40.56} & 31.41 & 46.15 & 77.58 & 31.13 & 50.77 & 57.91 & 52.72 \\

GPT-4  & $\ddagger$ & 30&75.68 & 72.73 & 86.12 & 40.42 & 31.60 & 45.99 & 78.36 & 32.40 & 52.10 & 62.18 & 53.65 \\

GPT-4  & $\S$& 30&74.72 & 72.21 & 85.13 & 40.30 & 33.95 & 46.17 & 78.04 & 31.79 & 53.28 & 60.25 & 53.61 \\

GPT-4 & $\P$  & 30& \textbf{76.29} & 70.88 & 85.79 & 40.34 & 32.43 & 45.79 & 76.71 & 33.00 & 52.76 & \textbf{60.39} & 53.40 \\
\midrule
%\midrule
%\multicolumn{12}{c}{\DeAR-Listwise (CoT Llama3.1 8B)}\\
%\midrule

Llama3.1-8B &$\dagger$ & 30 & 74.86 & 71.06 & 86.43 & 39.08 & 33.76 & 46.61 & 78.08 &\textbf{ 33.10 }& 53.17 & 59.55 & \textbf{53.72 }
%metrices_Llama-3.1-8B-Instruct-lora-8-cot-checkpoint-7000_part1
%metrices_Llama-3.1-8B-Instruct-lora-8-cot-checkpoint-7000_part2
\\  
Llama3.1-8B & $\ddagger$ & 30  & 75.54 & 70.39 & 86.53 & 38.48 & 34.32 & 46.20 & \textbf{78.47} & 31.69 & \textbf{53.79} & 59.43 & 53.61 
%metrices_Llama-3.1-8B-Instruct-lora-8-cot-checkpoint-7000
\\  
Llama3.1-8B & $\S$ & 30  & 75.29 & 71.17 & 88.10 & 39.05 & 31.47 & \textbf{46.69} & 77.77 & 32.64 & 53.09 & 57.93 & 53.35 
%metrices_Llama-3.1_32_64_0.1_50000_cot
\\  

%metrices_Llama-3.1_32_64_0.1_50000_cot
Llama3.1-8B & $\P$ & 30  & 75.33 & 70.02 & \textbf{88.36} & 38.80 & \textbf{35.04} & 46.34 & 77.34 & 32.56 & 52.24 & 58.63 & 53.66 
%metrices_Llama-3.1_32_64_0.1_50000_cot
\\

\bottomrule
\end{tabular} }

\caption{nDCG@10 performance on TREC Deep Learning Tracks (DL19, DL20) and BEIR datasets (CovidQA, NFCorpus, Touche, DBPedia, SciFact, Signal, News, Robust04). }
\label{table:benchmark}
\end{table*}

\subsection{Listwise Reranking with Reasoning}

The second stage of \DeAR performs listwise reranking on the top-$k$ candidates (e.g., $k=30$) from the pointwise stage. This stage enhances the ranking by reasoning over the candidate set, guided by synthetic reasoning chains generated by GPT-4o, addressing the limitations of pointwise methods in capturing inter-document dependencies.

\paragraph{Dataset Construction.} From MS MARCO~\citep{bajaj2016ms}, we sample 40K queries, as utilized by \citet{pradeep2023rankzephyr}. For each query $q_i$, we retrieve the top-20 candidate passages $\{d_{i1}, d_{i2}, \ldots, d_{i20}\}$ using GPT-4 via RankZephyr~\citep{pradeep2023rankzephyr,lin2021pyserini}. To generate teacher-labeled data, we employ GPT-4o with a CoT reasoning prompt, producing both a ranked list and corresponding reasoning for each query-passage set, as illustrated in Figure~\ref{fig:rankllm-train-example}.  
This process yields 20$K$ synthetic reasoning examples, each consisting of a query, candidate passages, CoT reasoning, and a teacher-generated ranking, which are used to train the student model. The prompt instructs the model to: (1) extract requirements, (2) match passages to them, and (3) rank documents using their IDs, as shown in Figure~\ref{fig:subfig-reason-prompt}. %Both the CoT and the final permutation are used as supervision targets,. 

\paragraph{Training Objective.} We train an instruction-tuned student model using supervised fine-tuning. For a query $q_i$ and its candidate set $D_i$, the input is the prompt containing $q_i$ and $D_i$, and the target output is the teacher-generated ranked list (e.g., \texttt{[1] > [2] > [3]}). Let $\pi_i = (\pi_{i1}, \pi_{i2}, \ldots, \pi_{ik})$ denote the target permutation, where $\pi_{ij} \in \{1, \ldots, k\}$ indicates the position of passage $d_{ij}$ in the ranked list (e.g., $\pi_{i1} = 2$ means $d_{i1}$ is ranked second). The student model generates a predicted permutation $\hat{\pi}_i$, which is optimized to align with the teacher’s ranking through supervised fine-tuning on the synthetic dataset.
% \paragraph{Training Objective.} We train an instruction-tuned student model using supervised fine-tuning. For a query $q_i$ and its candidate set $D_i$, the input is the prompt containing $q_i$ and $D_i$, and the target output is the teacher-generated ranked list (e.g., \texttt{[1] > [2] > [3]}). Let $\pi_i = (\pi_{i1}, \pi_{i2}, \ldots, \pi_{ik})$ denote the target permutation, where $\pi_{ij} \in \{1, \ldots, k\}$ indicates the position of passage $d_{ij}$ in the ranked list (e.g., $\pi_{i1} = 2$ means $d_{i1}$ is ranked second). The student model generates a predicted permutation $\hat{\pi}_i$. We optimise the model using a listwise loss based on the cross-entropy between the predicted and target permutations, treating the task as a sequence generation problem:
% \begin{equation}
% \mathcal{L}_{\text{listwise}}(\pi_i, \hat{\pi}_i) = -\frac{1}{k} \sum_{j=1}^k \log p(\hat{\pi}_{ij} = \pi_{ij} | q_i, D_i; \theta_s),
% \end{equation}
% where $p(\hat{\pi}_{ij} = \pi_{ij} | q_i, D_i; \theta_s)$ is the probability of predicting the correct passage identifier at position $j$, conditioned on the input prompt and model parameters $\theta_s$. 

\section{Experiments}

We evaluate \DeAR on standard reranking benchmarks and settings.% Our experiments aim to validate: (1) the effectiveness of pointwise and listwise stages across domains, (2) the impact of reasoning-augmented supervision, and (3) the generalizability across LLM families and sizes.

\subsection{Implementation Details}
\paragraph{Datasets and Evaluation Metrics.} We evaluate our approach on TREC-DL~\citep{bajaj2016ms}, BEIR~\citep{thakur2021beir}, and open-domain QA tasks such as NQ~\citep{kwiatkowski2019natural} and WebQA~\citep{berant-etal-2013-semantic}. For TREC, we use DL19 and DL20. From BEIR, we select eight diverse datasets: Covid, NFCorpus, Signal, News, Robust04, Touche, DBPedia, and SciFact.  Following standard reranking pipelines~\citep{nogueira2019multi,sun2023chatgpt}, we retrieve the top-100 candidate documents per query using BM25 via Pyserini and Rankify~\cite{Lin_etal_SIGIR2021_Pyserini,abdallah2025rankify}. We use NDCG@1, NDCG@5, and NDCG@10 to measure reranking performance, focusing on top-ranked relevance quality. We integrated our results with RankArena~\cite{abdallah2025rankarena} to show effective of our model compare with other reranker model. For open-domain QA, we also report top-1, top-10, and top-20 accuracy.

\paragraph{Models.} We experiment with both Qwen and LLaMA model families. For the \textbf{pointwise stage}, we use: (1) \textbf{Qwen:} Student models include Qwen 1.7B, 1.5B, 3B, 4B, and 7B, with Qwen3 14B as the teacher~\cite{bai2023qwen}. (2) \textbf{LLaMA:} Student models include LLaMA3.2 3B, 1B, and LLaMA3.1 8B, trained using LLaMA2 13B as teacher~\cite{touvron2023llama}. (3) For the \textbf{listwise stage}, we use a single student: LLaMA3.1 8B with reasoning supervision.

\begin{table}[!t]
\centering
\small
\setlength\tabcolsep{1pt}
\resizebox{0.50\textwidth}{!}{
\begin{tabular}{l | cc | ccc | c}

\toprule
Method& prev. & Top-$K$ & nDCG@1 & nDCG@5 & nDCG@10 & Avg \\

\midrule
BM25
& - & -& 33.33 & 45.96 & 55.77 & 45.02\\
\midrule
monoBERT (340M)& BM25 & 100 & 78.57 & 70.65 & 77.27 & 75.50\\
monoT5 (220M)& BM25 & 100& 83.33 & 77.46 & 81.27 & 80.69\\
monoT5 (3B)& BM25 & 100& 83.33 & 78.38 & 84.62 & 82.11\\
\midrule

RankGPT-3.5& BM25 & 100& 76.19 & 74.15 & 75.71 & 75.35\\

RankGPT-4& RankGPT-3.5& 20& 85.71 & 87.49 & 90.45 & 87.88\\
\midrule
\multicolumn{5}{c}{ \DeAR-Pointwise (\DeAR-P)} \\
\midrule

Llama3.1-8B (RL)$\dagger$ & BM25 & 100 &85.71  & 75.59 & 82.34 & 81.21
\\  
 Llama3.1-8B (BC)$\ddagger$ & BM25 & 100& 88.10  & 79.48 & 85.03 & 84.20
\\  
Llama3.1-3B (RL)$\S$& BM25 & 100 & 85.71 & 78.24 & 82.56 & 82.17
\\  
Llama3.1-3B (BC)$\P$ & BM25 & 100 & 85.71 & 76.94 & 82.12 & 81.59
\\

\midrule
\multicolumn{5}{c}{\DeAR-Listwise (\DeAR-L)}\\
\midrule

Llama3.1-8B &$\dagger$ & 30 
& 92.86  & 88.04 & 92.01 & 90.97
\\  

Llama3.1-8B & $\ddagger$ & 30 
& 92.86  & 88.04 & 90.98 & 90.63
\\  

Llama3.1-8B & $\S$ & 30 
& 90.48  & 90.32 & 92.05 & 90.95
\\  

Llama3.1-8B & $\P$ & 30
& 90.48  & 88.79 & 90.62 & 89.96
\\

\midrule
\end{tabular} }
\caption{Reranking results on NovelEval-2306. We compare BM25, monoT5, GPT baselines, and \DeAR (pointwise and listwise).}
\label{table:novel}
\end{table}

\paragraph{Training Infrastructure.} All experiments are conducted on 4\texttimes{}V100 GPUs (32GB). The teacher models remain frozen during distillation.  All models are fine-tuned using LoRA adapters. For the pointwise stage, we set \texttt{lora\_alpha} = 64. For the listwise stage, we reduce \texttt{lora\_alpha} to 8 to reflect reasoning efficiency. We use the Adam optimizer~\cite{kingma2014adam} with a batch size of 8, and a learning rate of 2e-5 with linear decay, weight decay of 0.1, and train for 3 epochs.

\subsection{Superior Performance}
\label{sec:superior-performance}
We evaluate \DeAR’s dual-stage reranking on TREC DL19/20 and BEIR datasets, using nDCG@10. Table~\ref{table:benchmark} presents full results, with key findings summarized below. \textbf{Strong Pointwise Performance.} In the pointwise stage, \DeAR-P re-ranks Top-100 BM25 candidates. LLaMA3.1-8B with RankLoss (RL) achieves a BEIR average of 52.09, improving +8.67 over BM25 (43.42) and surpassing monoT5-3B (51.36). Binary Cross-Entropy (BCE) excels on DL19 (74.50) and Touche (37.23), while RL outperforms overall (52.09 vs. 51.95). The compact LLaMA3.1-3B RL scores 50.59, confirming scalability. \textbf{Enhanced with Listwise CoT Reranking.} The listwise stage refines the Top-20 pointwise outputs using CoT reasoning. \DeAR-L (LLaMA3.1-8B RL) reaches 53.72 BEIR average, outperforming RankGPT-4 (53.68) on Covid (88.36 vs. 85.51), NFCorpus (40.56 vs. 38.47), and Robust04 (62.18 vs. 57.55), leveraging efficient open-source models. \textbf{Benefits of CoT Reasoning.} CoT enhances inter-document understanding, with LLaMA3.1-8B RL (53.72) surpassing GPT-4-BCE (53.65). It boosts weaker pointwise models, e.g., LLaMA3.1-3B-BCE improves from 50.58 to 53.61, closing performance gaps without increasing model size. \textbf{Scalability and Efficiency.} Both 8B and 3B models excel in listwise reranking. LLaMA3.1-3B RL achieves 88.10 on Covid and 53.35 BEIR average, nearing 8B performance. RL consistently outperforms BCE (53.77 vs. 53.67), optimizing permutations effectively.

\begin{table}[!t]
\centering
\small
\setlength\tabcolsep{2pt}
\resizebox{0.45\textwidth}{!}{
\begin{tabular}{l |ccc c}

\toprule
Method &  @1 & @5 & @10 & Avg \\

\midrule

BM25
& 54.26 & 52.78 & 50.58 & 52.54
\\

\midrule

RankGPT (text-davinci-003)
& 70.54 & 61.90 & 57.24 & 63.23
\\
RankGPT (gpt-3.5-turbo)
& 75.58 & 66.19 & 60.89 & 67.55
\\
RankGPT (gpt-4)
& 79.46 & 71.65 & 65.68 & 72.26
\\

\midrule
RankGPT  (rerank-english-v2.0)
& 79.46 & 71.56 & 64.78 & 71.27
\\

\midrule
RankGPT (claude-2)
& 66.66 & 59.33 & 55.91 & 60.63
\\
RankGPT (claude-instant-1)
& 81.01 & 66.71 & 62.23 & 69.98
\\

\midrule
RankGPT (text-bison-001)
& 69.77 & 64.46 & 58.67 & 64.30
\\
RankGPT (bard-2023.10.21)
& 81.01 & 65.57 & 60.11 & 68.90
\\

\midrule
RankGPT (flan-t5-xxl)
& 52.71 & 51.63 & 50.26 & 51.53
\\
RankGPT (ChatGLM-6B)
& 54.26 & 52.77 & 50.58 & 52.54
\\
RankGPT (Vicuna-13B)
& 54.26 & 51.55 & 49.08 & 51.63
\\

\midrule
\multicolumn{5}{c}{ \DeAR-Pointwise (\DeAR-P)} \\
\midrule

 Llama3.1-8B (RL)
& 80.23 & 69.86 & 64.16 & 71.42
\\  
Llama3.1-8B (BC)
& 79.46 & 71.45 & 65.48 & 72.13
\\  
Llama3.1-3B (RL)
& 78.68 & 70.19 & 64.26 & 71.04
\\  
Llama3.1-3B (BC)
& 83.33 & 72.30 & 65.43 & 73.69
\\

\midrule
\multicolumn{5}{c}{\DeAR-Listwise (\DeAR-L)}\\
\midrule

Llama3.1-8B
& 77.91 & 72.32 & 66.27 & 72.17
\\  
Llama3.1-8B
& 77.13 & 71.70 & 66.05 & 71.63
\\  
Llama3.1-8B
& 77.13 & 72.00 & 65.66 & 71.60
\\  
Llama3.1-8B
& 81.00 & 72.98 & 66.56 & 73.51
\\

\bottomrule
\end{tabular}}

\caption{Results of NDGC for different LLMs on re-ranking top-20 passages on DL-19.}

\label{table:more-llm}
%\vspace{-2mm}
\end{table}

\begin{figure*}[ht]
    \centering
    \includegraphics[width=0.9\linewidth]{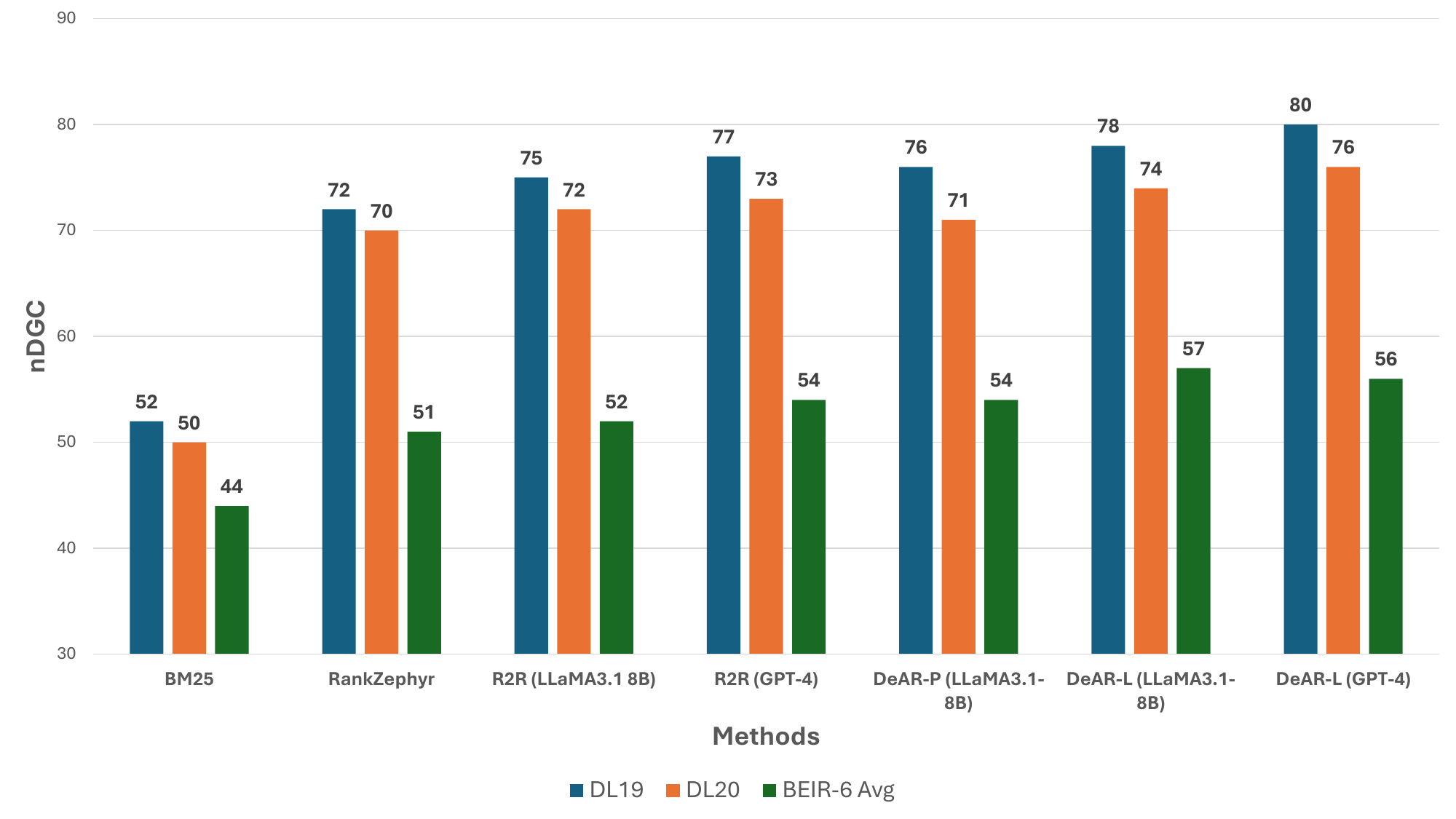}
\caption{nDCG@5 performance of \DeAR-Listwise vs. Reason-to-Rank (R2R) on TREC DL19/20 and BEIR datasets. See Appendix~\ref{sec:reasoning-comparison} and Table~\ref{tab:ndcg5_student_baseline} for full comparison.}
    \label{fig:reason-result}
\end{figure*}

\subsection{NovelEval Reranking Result}
\label{sec:novaleval-reranking-result}
NovelEval-2306 tests \DeAR on novel queries to ensure generalisation. In pointwise reranking, \DeAR-P (LLaMA3.1-8B BCE) scores 84.20 average across nDCG@1/5/10, outperforming monoT5-3B (82.11) and RankGPT-3.5 (75.35), and closely trailing RankGPT-4 (87.88). The 3B RL model achieves 82.17, underscoring \DeAR’s efficiency with smaller models. Listwise reranking with CoT, applied to the Top-20 pointwise outputs, significantly enhances performance: \DeAR-L (LLaMA3.1-8B RL) reaches 90.97 average, surpassing RankGPT-4 by +3.09, with nDCG@10 of 92.01. RL aids listwise optimization, as seen in the consistent gains over BCE (90.97 vs. 89.96). Table~\ref{table:novel} details these results, confirming \DeAR’s strong generalization on unseen queries using open-source models, avoiding reliance on proprietary APIs.

%To evaluate \DeAR’s robustness on truly novel queries, we test both pointwise and listwise reranking on the NovelEval-2306 benchmark, which features post-GPT-4 questions across four diverse domains to minimize memorisation and leakage. All pointwise variants outperform BM25 (45.02 avg), with the best model (LLaMA3.1-8B trained with Binary Cross-Entropy) achieving 84.20 on average, outperforming monoT5 and GPT-3.5 while remaining competitive with GPT-4. Smaller 3B models also perform strongly (e.g., 82.17 avg for RankLoss), affirming scalability. In the listwise stage, re-ranking the top-20 candidates with Chain-of-Thought reasoning yields further improvements: LLaMA3.1-8B-RL achieves 90.97 average, surpassing GPT-4 by 3.09 points, while even 3B variants reach 90.95. These gains are driven by CoT refinement and a strong pointwise foundation. RankLoss consistently benefits listwise optimization, highlighting the value of pairwise supervision. Overall, \DeAR demonstrates superior generalisation on unseen questions without reliance on proprietary APIs or large-scale models.

\subsection{ Performance on TREC DL-19}
\label{sec:dl19-llm-reranking}
On TREC DL-19, a standard IR benchmark, \DeAR competes with proprietary LLM APIs like GPT-4 and Claude. In the pointwise stage, \DeAR-P (LLaMA3.1-3B BCE) re-ranks Top-100 BM25 candidates, achieving 73.69 average (nDCG@1/5/10), outperforming RankGPT-4 (72.26) and other APIs like Claude-2 (60.63) and Bard (68.90). LLaMA3.1-8B RL scores 71.42, remaining competitive. The listwise stage, re-ranking the Top-20 pointwise outputs with CoT, further improves performance: \DeAR-L (LLaMA3.1-8B) reaches 73.51, exceeding RankGPT-4 and demonstrating superior refinement. Table~\ref{table:more-llm} presents these results, validating \DeAR’s state-of-the-art performance with compact, open-source models in real-world IR settings.

%To assess how \DeAR compares against widely-used proprietary APIs, we evaluate its reranking performance on the TREC DL-19 benchmark, a standard testbed for deep learning-based retrieval. This experiment allows a controlled comparison against LLM APIs such as GPT-4, Claude, PaLM, and Bard, which are commonly used for reranking via prompt engineering. We re-rank the top-20 BM25 candidates and report nDCG@1, @5, and @10 (Table~\ref{table:more-llm}). \DeAR-Pointwise, particularly the LLaMA3.1-3B model trained with Binary Cross-Entropy, outperforms GPT-4 (73.69 vs. 72.26 average) while using a smaller open-source model. Other variants remain competitive, often surpassing GPT-3.5 and Claude. The listwise stage further improves ranking through Chain-of-Thought refinement, with LLaMA3.1 8B achieving 73.51, again exceeding GPT-4. These results demonstrate that \DeAR delivers state-of-the-art reranking performance without relying on commercial APIs, validating its scalability and effectiveness in real-world IR settings.

\begin{table}[ht]
\setlength\tabcolsep{2pt}
\centering
\scriptsize

\begin{tabular}{ll|ccc}

    \toprule
   Teacher  & Student  & dl19&	dl20& Avg BEIR \\
    \midrule

   Qwen3-14B  & Qwen3-1.7B  & 73.31	&68.00		&49.58 
       %rankLLaMA_teacher_Qwen3-14B_student_Qwen3-1.7B_msmacroo_0.1_2/checkpoint-1000
   \\
    Qwen3-14B  & Qwen3-4B  & 74.04&	66.94&	50.59
    %rankLLaMA_teacher_Qwen3-14B_student_Qwen3-4B_msmacroo_0.1_2
    \\

   Qwen2.5-14B  & Qwen2.5-7B  &  74.06	&  66.15	&  51.24
   %rankLLaMA_teacher_Qwen2.5-14B_student_Qwen2.5-7B_msmacroo_0.1_2/checkpoint-1400
   \\

   Qwen2.5-14B  & Qwen1.5-1.5B  & 73.63 &	65.85	&50.00
   %rankLLaMA_teacher_Qwen2.5-14B_student_Qwen2.5-1.5B_msmacroo_0.1_2/checkpoint-1000
   \\
    Qwen2.5-14B  & Qwen1.5-3B  & 73.47&	66.42	&	50.20
   %rankLLaMA_teacher_Qwen2.5-14B_student_Qwen2.5-3B_msmacroo_0.1_2/checkpoint-800
   \\
\midrule
   LLaMA2-13B  & LLaMA-3.2-3B  &  74.49& 	69.02& 	50.58
%rankLLaMA_teacher_13b_student_LLaMA-3.2-3B_msmacroo_0.1_2/checkpoint-11400
\\
LLaMA2-13B  & LLaMA-3.2-1B  & 	72.82&	68.08	&49.72

%%rankLLaMA_teacher_13b_student_1b_msmacroo_0.1_2/checkpoint-1200	
\\

	LLaMA2-13B  & LLaMA-3.1-8B  & 74.50	&  68.71	& 51.95

    %LLaMA8_passage2	rankLLaMA_teacher_13b_student_8b_msmacroo_0.1_2	checkpoint-1000
\\
    \bottomrule
    %%%%%%%%% all these from leonardo server 
\end{tabular}
\caption{nDCG@10 performance (in percentage) of \DeAR-Pointwise with different teacher and student model pairs using binary cross-entropy across TREC and BEIR datasets. See Appendix~\ref{app:ablation-teacher-student} and Table~\ref{tab:diffmodels_app} for full comparison.}
\label{tab:diffmodels}
\end{table}

\subsection{Comparison with Reasoning Methods}
\label{sec:reasoning-comparison}
We compare \DeAR’s CoT reasoning with Reason-to-Rank (R2R)~\citep{ji2024reasoningrank}, which employs direct relevance and comparison reasoning, using nDCG@5 on TREC DL19/20 and BEIR datasets. In the pointwise stage, \DeAR-P (LLaMA3.1-8B RL) provides a strong foundation, e.g., 89.13 on Covid. Listwise reranking with CoT, applied to the Top-20 pointwise outputs, significantly outperforms R2R: \DeAR-L (LLaMA3.1-8B BCE) achieves 80.71 on DL19, surpassing R2R-GPT-4 (77.7) and R2R-LLaMA3.1-8B (75.4). On BEIR, \DeAR-L scores 91.94 on Covid and 69.49 on Robust04, outpacing R2R-GPT-4 (85.3 and 58.6). The 3B RL model reaches 91.94 on Covid, rivaling R2R’s larger models. On NFCorpus, \DeAR’s 45.76 exceeds the result by R2R-LLaMA3.1-8B (36.4). Figure~\ref{fig:reason-result} visualizes these gains, emphasizing \DeAR’s robust multi-document reasoning with open-source efficiency, enhancing pointwise outputs through CoT-guided global ranking decisions.

\section{Additional Analysis}

\subsection{Impact of Teacher–Student}
\label{sec:ablation-teacher-student}
We assess \DeAR-P’s pointwise reranking with various teacher-student pairs. LLaMA2-13B to LLaMA3.1-8B yields the highest BEIR average (51.95), while Qwen2.5-14B to Qwen2.5-7B scores 51.24. Smaller students, like LLaMA-3.2-1B (49.72) and Qwen3-1.7B (49.58), remain competitive, showcasing \DeAR’s flexibility. Listwise reranking, applied to the Top-20 pointwise outputs, was not evaluated here but is expected to further enhance these results, as seen in prior subsections. Table~\ref{tab:diffmodels} details the pointwise performance, highlighting \DeAR’s adaptability across diverse model architectures and sizes for efficient knowledge distillation.

\subsection{Alpha Selection for KL Divergence}
\label{sec:ablation-alpha}
\begin{figure}[t]
    \centering
    \includegraphics[width=0.9\linewidth]{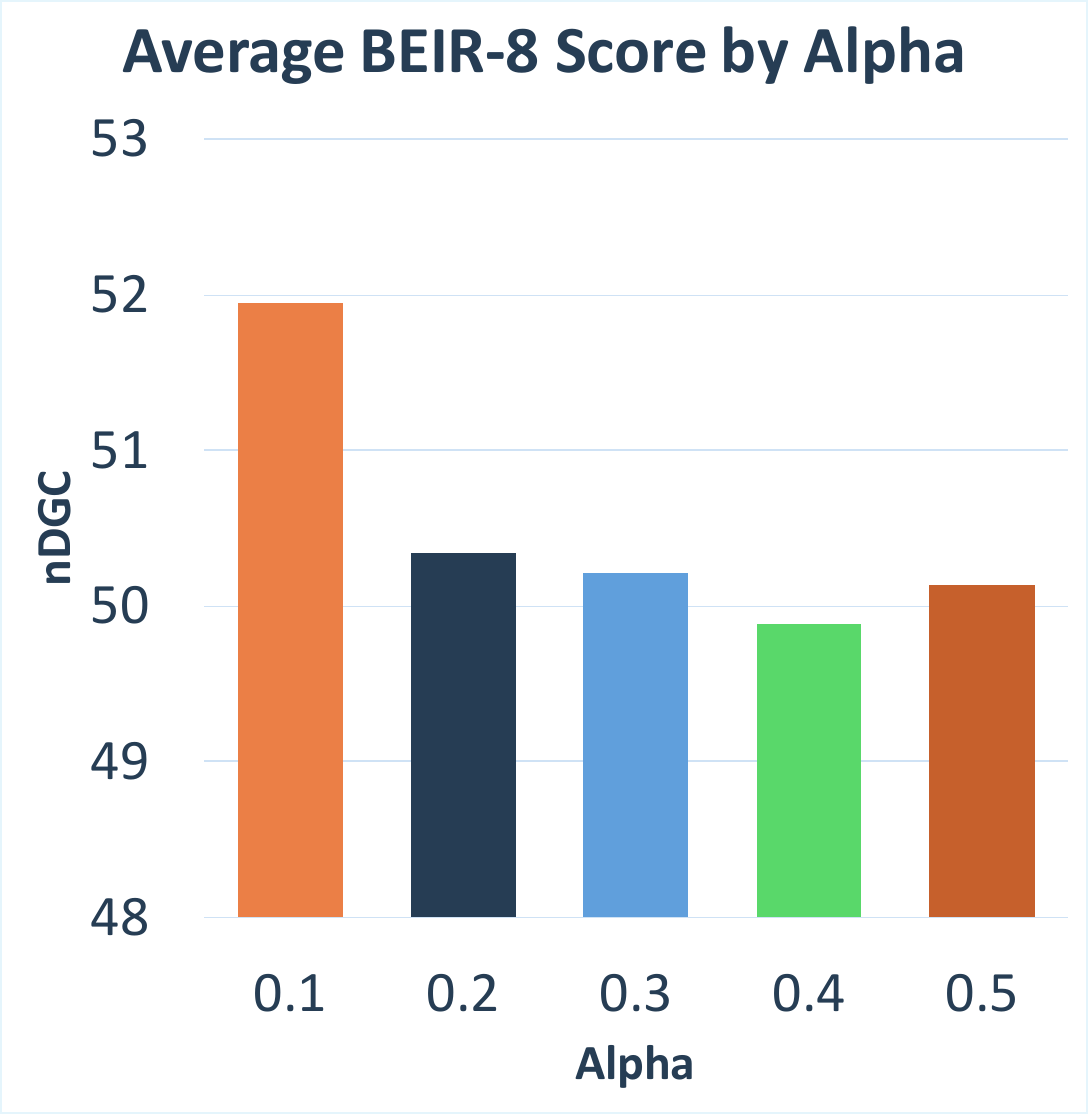}
    \caption{Average BEIR-8 score (nDCG@10, in percentage) across alpha values for \DeAR-Pointwise (Teacher: LLaMA2-13B, Student: LLaMA-3.1-8B) with KL divergence and binary cross-entropy loss.}
    \label{fig:alpha-beir}
\end{figure}

We analyze now the impact of the alpha coefficient, which balances KL divergence and binary cross-entropy (BCE) during pointwise reranking in \DeAR, using the LLaMA2-13B $\rightarrow$ LLaMA-3.1-8B teacher-student pair. Varying $\alpha$ from 0.1 to 0.5 on MS MARCO, we evaluate nDCG@10 across eight BEIR datasets. Figure~\ref{fig:alpha-beir} shows the average BEIR score peaking at 52.5 with $\alpha=0.1$, then declining to 48.5 at $\alpha=0.5$. Lower $\alpha$ prioritizes the BCE ranking objective, optimizing relevance, while higher $\alpha$ overemphasizes teacher logit alignment, reducing ranking quality. Thus, we adopt $\alpha=0.1$ for all KL-based distillation in \DeAR.

\subsection{Inference Time vs.  Performance}
\label{sec:time-ndcg}
\begin{figure}
    \centering
    \includegraphics[width=\linewidth]{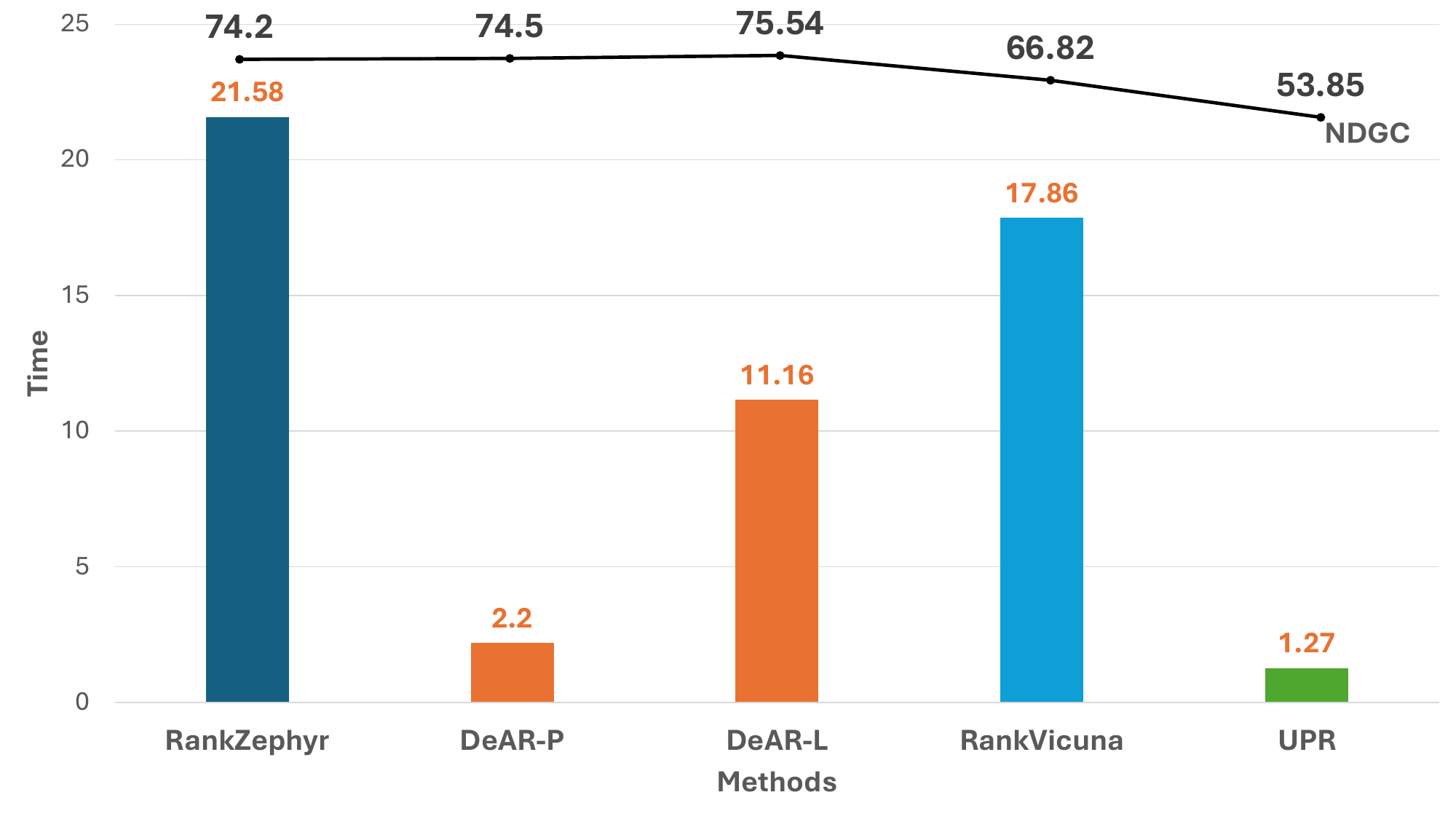}
    \caption{nDCG@10 performance vs. inference time (seconds) Per Query for \DeAR-P, \DeAR-L, RankZephyr, RankVicuna, and UPR on TREC DL19.}
    \label{fig:time-ndcg}
\end{figure}

We next compare \DeAR’s inference time and nDCG@10 on TREC DL19 (Figure~\ref{fig:time-ndcg}). \DeAR-P (LLaMA3.1-8B) achieves 74.5 nDCG in 2.2s, outperforming RankZephyr (74.2, 21.58s) and RankVicuna (66.82, 17.86s), with only UPR faster (1.27s, 53.85 nDCG). \DeAR-L (LLaMA3.1-8B) reaches 75.54 nDCG in 11.16s, balancing speed and CoT-enhanced accuracy, surpassing slower baselines with open-source efficiency.
%We compare the inference time and nDCG@10 performance of \DeAR against baselines on TREC DL19, as shown in Figure~\ref{fig:time-ndcg}. In the pointwise stage, \DeAR-P (LLaMA3.1-8B) achieves a competitive nDCG of 74.5 with the second-fastest inference time of 2.2 seconds, surpassing RankZephyr (74.2 nDCG, 21.58s) and RankVicuna (66.82 nDCG, 17.86s). UPR is fastest at 1.27s but scores only 53.85 nDCG. In the listwise stage, \DeAR-L (LLaMA3.1-8B) attains the highest nDCG of 75.54 with a moderate 11.16s, balancing efficiency and superior ranking quality. This demonstrates \DeAR-P’s speed advantage for pointwise reranking and \DeAR-L’s enhanced accuracy via CoT reasoning, both outperforming slower baselines while maintaining open-source efficiency.
\label{sec:time-ndgc}

\begin{table*}[!ht]

\setlength{\tabcolsep}{2pt}

%\addtolength{\tabcolsep}{-0.65pt}
\small
\centering
\resizebox{0.9\textwidth}{!}{
\begin{tabular}{@{}l |l| c c c  | c c c@{}}\toprule
Reranking/ &  Model  &  \multicolumn{3}{c}{NQ} & \multicolumn{3}{c}{WebQ}  \\

& & Top-1 & Top-10 & Top-50 & Top-1 & Top-10 & Top-50   \\
\midrule
BM25      & - & 23.46 & 56.32 & 74.57 & 19.54  & 53.44 & 72.34 \\

\midrule
 
\multirow{2}{*}{UPR~\cite{sachan2022improving}}  %& T5-small& 23.60 & 59.97  & 75.15 &  18.55 &  55.56   &  72.68  \\

  %& T5-base & 26.81 & 63.32 & 76.12 & 20.66 & 58.56 & 72.68  \\
  %& T5-large& 29.75 & 65.67 & 76.48 & 24.21 & 60.38 & 73.32 \\
  & T0-3B & 35.42 & 67.56 & 76.75 & 32.48 & 64.17 & 73.67 \\
  %& gpt2 & 25.95 & 60.47 & 75.87 & 20.47 & 56.49 & 72.78 \\
 % & gpt2-medium &26.75 & 63.04 & 75.95 & 22.39 & 59.54 & 72.44 \\
 % &gpt2-large & 26.59 &  62.68  & 75.95 & 24.06   &  59.84  & 72.78  \\

%  & gpt2-xl & 27.28  & 63.24 & 75.84 &  23.57   & 60.48 &  72.73  \\
  &gpt-neo-2.7B &28.75  & 64.81  &  76.56 &24.75    & 59.64 &  72.63  \\

\midrule

 \multirow{1}{*}{RankGPT~\cite{sun2023chatgpt}} & LLaMAv3.1-8b & 41.55 & 66.17 & 75.42 & 38.77 & 62.69 & 73.12 \\

\midrule
  
  \multirow{1}{*}{RankT5~\cite{zhuang2023rankt5}}
 % & base & 43.04 & 68.47 & 76.28 & 36.95 & 64.27 & 74.45  \\
 % & large &45.54 & 70.02 & 76.81 & 38.77 & 66.48 & 74.31 \\
  &3b & 47.17& 70.85 & 76.89 & 40.40 & 66.58 & 74.45  \\

\midrule

\multirow{1}{*}{Inranker~\cite{laitz2024inranker}}  
%& small & 15.90 & 46.84 & 69.83  & 14.46  & 46.25 & 69.98 \\
%&base  &15.90 & 48.11 & 69.66 &14.46 & 46.80 & 69.68 \\
&3b &15.90 & 48.06 & 69.00 & 14.46 & 46.11 & 69.34 \\
\midrule

\multirow{1}{*}{MonoBert~\cite{monobert}} 
& large & 39.05 & 67.89 & 76.56 & 34.99 &64.56 & 73.96 \\

\midrule

\multirow{1}{*}{Twolar~\cite{baldelli2024twolar}} 
& twolar-xl & 46.84 & 70.22 & 76.86 &41.68 &67.07 &74.40 \\

\midrule

\multirow{1}{*}{Echorank~\cite{rashid2024ecorank}} 
%& flan-t5-large& 36.73 & 59.11 & 62.38 & 31.74 &58.75 &61.51 \\
&flan-t5-xl&   41.68 & 59.05 & 62.38  & 36.22 &57.18 &61.51\\

\midrule

\multirow{2}{*}{\makecell[l]{Incontext\\ Reranker~\cite{chen2024attentionlargelanguagemodels}}} &  \multirow{2}{*}{LLaMAv3.1-8b} & \multirow{2}{*}{15.15} & \multirow{2}{*}{57.11}  & \multirow{2}{*}{76.48}  & \multirow{2}{*}{18.89}   & \multirow{2}{*}{52.16} &  \multirow{2}{*}{71.70} \\
&&&&&&&\\

\midrule

\multirow{1}{*}{Lit5~\cite{tamber2023scaling}}
%&  LiT5-Distill-base  & 40.05 & 65.95 & 75.73  & 36.76 & 63.48 & 73.12 \\
%&  LiT5-Distill-large  & 44.40 & 67.59 & 76.01   &  39.66  &64.56 &73.67\\
%&  LiT5-Distill-xl  &  47.81 & 68.55 & 76.26  &  42.37  &65.55 &73.62 \\
%& LiT5-Distill-base-v2   &  42.57 &  66.73 & 75.56  & 39.61 &64.22 &73.32\\
%&  LiT5-Distill-large-v2  & 46.53 &  67.83 & 75.87  & 41.97 &65.64 &72.98 \\
&  LiT5-Distill-xl-v2  & 47.92 & 69.03 & 76.17  & 41.53 &65.69 &73.27  \\

\midrule

\multirow{2}{*}{ \makecell[l]{Sentence \\ Transformer } }
%& GTR-base & 39.41 &65.95 & 76.03  & 36.56 &64.32 &73.62  \\
%& GTR-large & 40.63 &68.25 & 76.73  & 38.97 &65.30 &73.57 \\
%& T5-base & 31.19 &63.60 & 76.06  & 29.77 &62.84 &73.52 \\
%& T5-large & 30.80 &63.35 &76.37   & 30.51  &61.71 &73.37 \\
%&   all-MiniLM-L6-v2 & 33.35  & 65.37  &  76.01 & 30.95   & 62.10  & 73.52 \\
%& GTR-xl   & 41.55  & 67.78  & 76.81 &  38.92   &  66.04  & 74.01 \\
& GTR-xxl   & 42.93 &  68.55&  77.00 &39.41    & 65.89 & 74.01 \\
&  T5-xxl  & 38.89  &  67.78  &76.64   &  35.82  & 65.20  & 74.01 \\
%&  Bert-co-condensor  & 30.96  & 61.91   &  75.20 &  32.43  &  62.20  & 73.08 \\
%&  Roberta-base-v2  & 32.60  & 63.24  & 75.42  &  31.34  &  62.64 &  73.37 \\
\midrule

\DeAR-P & Llama3.1 8B
& 48.92  &73.35 & 78.78 & 41.93  &    67.67 & 75.05 \\

\DeAR-L & Llama3.1 8B
& 54.29  &73.07 & 78.78 & 46.60  &    68.11 & 75.05 \\

\bottomrule           
\end{tabular}}
\caption{Performance of re-ranking methods on BM25-retrieved documents for NQ Test and WebQ Test. Results are reported in terms of Top-1, Top-10, and Top-50 accuracy. Note that some results (e.g., UPR) differ from original papers due to re-ranking top-100 documents instead of 1,000.
}
\label{tab:open-domain-results}
\end{table*}
\subsection{Open Domain QA}
\label{sec:open-domain-qa}
We finally evaluate \DeAR’s generalization to Natural Questions (NQ) and Web Questions (WebQ) without Wikipedia fine-tuning, re-ranking Top-100 BM25 passages (Table~\ref{tab:open-domain-results}). \DeAR-P (LLaMA3.1-8B, RankLoss) scores 48.92 Top-1 on NQ and 41.93 on WebQ, outperforming RankGPT (41.55, 38.77) and Twolar (46.84, 41.68). \DeAR-L with CoT boosts Top-1 to 54.29 (NQ) and 46.60 (WebQ), surpassing RankT5-3B (47.17, 40.40). Top-10 (73.07 NQ, 68.11 WebQ) and Top-50 (78.78 NQ, 75.05 WebQ) gains are smaller, but CoT enhances precision under domain shift, excelling with open-source models trained on MS MARCO.
%We assess \DeAR’s generalization to out-of-domain open QA datasets—Natural Questions (NQ) and Web Questions (WebQ)—without fine-tuning on Wikipedia, re-ranking Top-100 BM25 passages. Table~\ref{tab:open-domain-results} reports Top-1, Top-10, and Top-50 accuracy. In the pointwise stage, \DeAR-P (LLaMA3.1-8B, RankLoss) achieves 48.92 Top-1 on NQ and 41.93 on WebQ, surpassing RankGPT (41.55, 38.77), Twolar (46.84, 41.68), and MonoBERT (39.05, 34.99). Listwise reranking with CoT (\DeAR-L, LLaMA3.1-8B RL $\rightarrow$ LLaMA3.1-8B) boosts Top-1 to 54.29 on NQ and 46.60 on WebQ, outperforming RankT5-3B (47.17, 40.40) and Lit5 (47.92, 41.53). Top-10 (73.07 NQ, 68.11 WebQ) and Top-50 (78.78 NQ, 75.05 WebQ) show modest gains, but the Top-1 improvement highlights CoT’s precision under domain shift. \DeAR’s dual-stage approach, trained only on MS MARCO, excels in unseen QA tasks, leveraging open-source efficiency.

\section{Conclusion}
\label{sec:conclusion}

\textbf{DeAR} introduces a dual-stage reranking framework that decouples pointwise scoring and listwise reasoning, achieving high accuracy and interpretability. Stage 1 distills relevance signals from a 13B LLaMA teacher into {3, 8}B students using hybrid losses, ensuring robust calibration. Stage 2 fine-tunes with GPT-4o-generated CoT permutations for global reasoning. \DeAR achieves 90.97 nDCG@10 on NovelEval, surpassing GPT-4 by +3.09, and 54.29 Top-1 accuracy on Natural Questions, outperforming MonoT5 and RankGPT. With an inference time of 2.2s (pointwise) and 11.16s (listwise), \DeAR offers an efficient, open-source solution for advanced reranking.

\section*{Limitations}
\label{sec:limitations}

While \DeAR achieves state-of-the-art performance in document reranking, it has several limitations. First, the dual-stage training pipeline relies on synthetic data generated by GPT-4o for listwise reasoning, which may introduce biases or errors from the teacher model, such as hallucinations or misrankings, potentially affecting generalization. Second, the framework's performance is evaluated on top-100 candidates retrieved by BM25, making it dependent on the quality of the initial retrieval stage. Third, the listwise stage processes smaller candidate sets (e.g., top-20), which may limit its ability to handle larger sets due to context window constraints in LLMs. Finally, while \DeAR is efficient compared to baselines like RankZephyr, the two-stage process increases computational complexity compared to single-stage rerankers, which may pose challenges for resource-constrained environments.
% Bibliography entries for the entire Anthology, followed by custom entries
%\bibliography{anthology,custom}
% Custom bibliography entries only
\bibliography{custom}

\appendix

%\section{Appendix}
%\label{sec:appendix}
\section{Synthetic Data Generation Prompt}
\label{app:Synthetic}
\begin{figure}[t]
    \centering
    \includegraphics[width=0.5\textwidth]{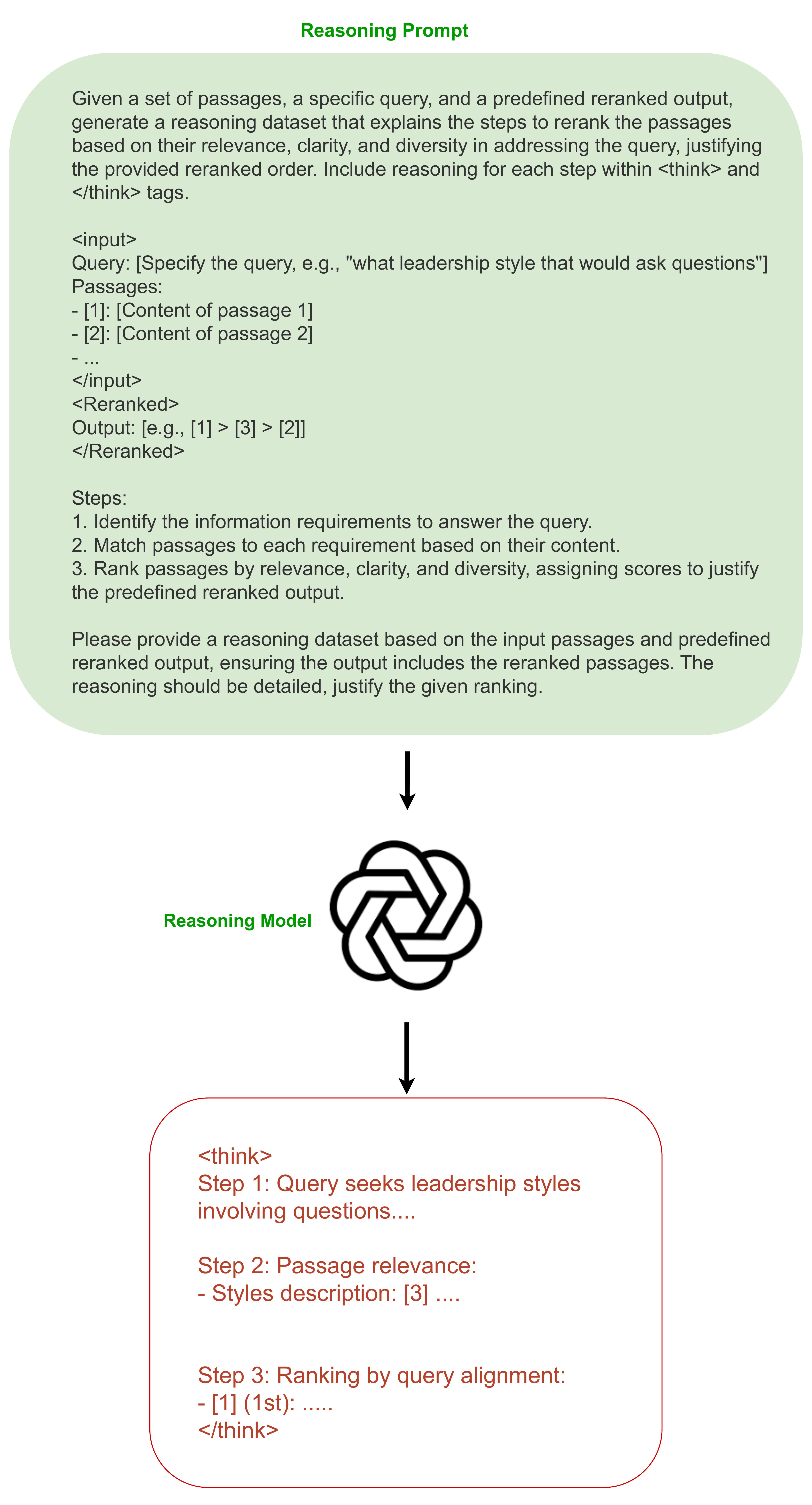}
    \caption{Prompt used to generate synthetic chains.}
    \label{fig:subfig-reason-prompt}
\end{figure}

Figure~\ref{fig:subfig-reason-prompt} illustrates the prompt used to generate synthetic chain-of-thought (CoT) reasoning examples for the listwise reranking stage of \DeAR. This prompt, provided to GPT-4o, instructs the model to: (1) identify information requirements for a given query, (2) match candidate passages to these requirements, and (3) produce a ranked list of passage identifiers with step-by-step reasoning. The resulting 20K synthetic examples, each comprising a query, candidate passages, CoT reasoning, and ranked output, are used to fine-tune the student model (LLaMA3.1-8B) for listwise reranking, enhancing its ability to reason globally over document sets and generate interpretable rankings.

\section{Comparison with Reasoning Models}
\label{app:reasoning-comparison}
We compare \DeAR’s chain-of-thought (CoT) reasoning approach with Reason-to-Rank (R2R)~\citep{ji2024reasoningrank}, which uses direct relevance and comparative reasoning, alongside other baselines. Table~\ref{tab:ndcg5_student_baseline} reports nDCG@5 performance across TREC DL19/20 and six BEIR datasets (Covid, Touche, News, NFCorpus, Robust04, DBPedia). Figure~\ref{fig:reason-result} visualizes these results, highlighting \DeAR’s gains.

In the pointwise stage, \DeAR-P (LLaMA3.1-8B BC) achieves 76.52 on DL19 and 54.65 BEIR average, outperforming R2R-LLaMA3.1-8B (75.4 and 52.0) and competitive with R2R-GPT-4 (77.7 and 54.4). The 3B RL model scores 53.39 BEIR average, showcasing scalability. In the listwise stage, \DeAR-L (LLaMA3.1-8B BC with CoT GPT) reaches 80.71 on DL19 and 56.92 BEIR average, surpassing R2R-GPT-4 (77.7 and 54.4) and R2R-LLaMA3.1-8B (75.4 and 52.0). Notable gains occur on Covid (90.53 vs. 85.3), Robust04 (69.02 vs. 58.6), and NFCorpus (45.75 vs. 36.3). \DeAR-L with CoT LLaMA achieves 57.78 BEIR average, with strong performance on Covid (91.44) and Robust04 (65.69), demonstrating robust multi-document reasoning.

These results, visualized in Figure~\ref{fig:reason-result}, confirm \DeAR’s CoT-guided listwise reranking enhances performance over R2R’s reasoning strategies, leveraging open-source efficiency to outperform larger proprietary models while maintaining interpretability.
\begin{table*}[ht]
\centering
\caption{NDCG@5 performance (in percentage) for student models and baseline comparisons across multiple datasets.}
\scriptsize
\setlength{\tabcolsep}{1pt}
\renewcommand{\arraystretch}{1.15}
\resizebox{0.8\textwidth}{!}{
\begin{tabular}{l|cc|ccc|cccccc}
\toprule
\textbf{Models}& prev. & Top-$K$&  \textbf{DL19} & \textbf{DL20} & \textbf{BEIR-6 Avg.} & \textbf{Covid} & \textbf{Touche} & \textbf{News} & \textbf{NFCorpus} & \textbf{Robust04} & \textbf{DBPedia} \\
\midrule
\multicolumn{10}{c}{\textbf{Baseline Models}} \\ \midrule
BM25 & -& -&52.78 & 50.67 & 44.04 & 63.24 & 48.11 & 41.28 & 35.66 & 43.59 & 32.36 \\ \midrule
DeBERTa & BM25 & 100& 68.5 & 64.2 & 47.3 & 73.4 & 32.1 & 50.2 & 33.7 & 49.2 & 45.4 \\
MonoT5& BM25 & 100 & 74.5 & 70.4 & -- & 80.0 & 34.1 & -- & -- & 46.0 & 35.2 \\ 
RankVicuna& BM25 & 100 & 71.1 & 68.7 & - & 67.1 & 48.7 & -- & 38.5 & 55.7 & 35.3 \\ 
RankZephyr & BM25 & 100& 72.2 & 70.5 & 51.7 & 85.1 & 36.5 & 53.3 & 38.9 & 60.7 & 35.5 \\ 
APEER & BM25 & 100& 74.6 & 72.3 & 51.1 & 83.9 & 35.3 & 52.1 & 33.4 & 56.0 & 46.1 \\
R2R (GPT-4) & BM25 & 100& 77.7 & 73.2 & 54.4 & 85.3 & 38.3 & 58.4 & 36.3 & 58.6 & 49.5 \\ 
R2R (Claude)& BM25 & 100 & 72.1 & 70.0 & 51.8 & 84.0 & 37.3 & 55.5 & 36.4 & 52.7 & 44.9 \\ 
R2R (Gemini)& BM25 & 100 & 71.4 & 68.6 & 51.0 & 83.5 & 37.0 & 53.8 & 36.1 & 51.8 & 43.7 \\ 
R2R (LLaMA3.1 8B)& BM25 & 100 & 75.4 & 72.4 & 52.0 & 84.6 & 36.2 & 53.8 & 36.4 & 53.5 & 47.9 \\
\midrule
\multicolumn{9}{c}{ \DeAR-Pointwise (\DeAR-P)} \\
\midrule
 Llama3.1-8B (RL)$\dagger$& BM25 & 100 & 75.86 & 70.41 & 54.93 & 89.13 & 38.73 & 52.17 & 41.19 & 60.43 & 47.90 \\
 Llama3.1-8B (BC)$\ddagger$& BM25 & 100 & 76.52 & 71.85 & 54.65 & 87.11 & 39.47 & 53.40 & 41.42 & 57.95 & 48.53 \\
Llama3.1-3B (RL)$\S$ & BM25 & 100 & 76.08 & 72.64 & 53.39 & 86.06 & 37.58 & 51.33 & 40.70 & 56.06 & 48.58 \\
Llama3.1-3B (BC)$\P$ & BM25 & 100 & 78.15 & 72.87 & 53.15 & 86.08 & 39.63 & 49.54 & 40.08 & 56.61 & 46.96 \\
\midrule
\multicolumn{10}{c}{ \DeAR-Listwise (CoT GPT)} \\ \midrule
GPT-4 &$\dagger$ & 30& 78.88 & 76.18 & 56.26 & 90.29 & 32.44 & 51.74 & 45.52 & 69.49 & 48.07 \\
GPT-4  & $\ddagger$ & 30& 80.71 & 76.11 & 56.92 & 90.53 & 34.66 & 52.89 & 45.75 & 69.02 & 48.69 \\
GPT-4  & $\S$& 30& 78.09 & 76.77 & 57.33 & 89.36 & 37.67 & 53.42 & 45.76 & 69.43 & 48.35 \\
GPT-4 & $\P$  & 30& 80.09 & 74.98 & 56.99 & 89.53 & 37.65 & 52.53 & 45.54 & 68.89 & 47.79 \\
% \midrule
% \multicolumn{10}{c}{ \DeAR-Listwise (LLaMA)} \\ \midrule
% \code{P-LLaMA3.1-8B-RL \rightarrow LLaMA3.1 8B} & 75.98 & 75.49 & 59.04 & 90.53 & 36.34 & 54.44 & 43.69 & 66.35 & 49.99 \\
% \code{P-LLaMA3.1-8B-BC \rightarrow LLaMA3.1 8B} & 75.23 & 74.08 & 58.16 & 89.48 & 36.27 & 53.07 & 43.41 & 67.15 & 49.61 \\
% \code{P-LLaMA3.1-3B-RL \rightarrow LLaMA3.1 8B} & 77.88 & 75.14 & 58.79 & 90.99 & 36.50 & 54.07 & 42.91 & 66.22 & 49.84 \\
% \code{P-LLaMA3.1-3B-BC \rightarrow LLaMA3.1 8B} & 77.00 & 74.72 & 57.68 & 90.62 & 37.78 & 52.24 & 42.82 & 66.19 & 49.37 \\
\midrule
\multicolumn{10}{c}{ \DeAR-Listwise (CoT LLaMA)} \\ \midrule
Llama3.1-8B &$\dagger$ & 30 &  76.93 & 75.63 & 56.18 & 88.49 & 37.41 & 52.08 & 43.31 & 66.45 & 49.33 \\
Llama3.1-8B & $\ddagger$ & 30  & 78.23 & 74.27 & 57.12 & 89.91 & 38.94 & 55.10 & 42.93 & 66.49 & 49.36 \\
Llama3.1-8B & $\S$ & 30 &76.91 & 75.40 & 56.69 & 91.94 & 35.73 & 52.68 & 43.92 & 66.26 & 49.63 \\
Llama3.1-8B & $\P$ & 30  & 78.40 & 74.28 & 57.78 & 91.44 & 38.03 & 51.85 & 43.81 & 65.69 & 49.86 \\
\bottomrule
\end{tabular}
}
\label{tab:ndcg5_student_baseline}
\end{table*}

\section{Impact of Teacher–Student Pairing}
\label{app:ablation-teacher-student}
We extend the ablation study from Section~\ref{sec:ablation-teacher-student} to evaluate \DeAR-Pointwise’s performance across various teacher–student pairs using binary cross-entropy loss, as shown in Table~\ref{tab:diffmodels_app}. The table reports nDCG@10 across TREC DL19/20 and eight BEIR datasets (Covid, DBPedia, News, NFCorpus, Robust04, Scifact, Signal, Touche). The LLaMA2-13B to LLaMA3.1-8B pairing achieves the highest BEIR average (51.95), with strong performance on Covid (84.14) and Robust04 (52.43). Qwen2.5-14B to Qwen2.5-7B follows closely with a BEIR average of 51.24, excelling on Scifact (76.52) and DBPedia (46.30). Smaller models remain competitive: LLaMA-3.2-1B (49.72 BEIR average) and Qwen3-1.7B (49.58) perform robustly, particularly on Touche (37.18 and 31.91, respectively). These results, visualized for the LLaMA2-13B to LLaMA3.1-8B pair in Figure~\ref{fig:alpha-beir} for alpha selection, highlight \DeAR’s flexibility across diverse model architectures and sizes. Listwise reranking, applied to the top-20 pointwise outputs, is expected to further enhance these results, as demonstrated in Section~\ref{sec:superior-performance}.

\begin{table*}[!ht]
\setlength\tabcolsep{2pt}
\centering
\scriptsize

\begin{tabular}{ll|cccccccccc|c}

    \toprule
   Teacher  & Student  & dl19&	dl20	&covid&	dbpedia	&news	&nfc	&robust04	&scifact	&signal&	touche & Avg BEIR \\
    \midrule

   Qwen3-14B  & Qwen3-1.7B  & 73.31	&68.00	&83.25	&43.69	&49.25	&35.70	&49.96	&74.88&	28.07&	31.91	&49.58 
       %rankLLaMA_teacher_Qwen3-14B_student_Qwen3-1.7B_msmacroo_0.1_2/checkpoint-1000
   \\
    Qwen3-14B  & Qwen3-4B  & 74.04&	66.94&	82.41&	45.47&	50.92&	35.56&	52.07&	76.09&	28.57&	33.68&	50.59
    %rankLLaMA_teacher_Qwen3-14B_student_Qwen3-4B_msmacroo_0.1_2
    \\

   Qwen2.5-14B  & Qwen2.5-7B  &  74.06	&  66.15	&  83.21	& 46.30	& 49.88	& 35.83	& 52.86& 	76.52	& 29.09	&36.29	&  51.24
   %rankLLaMA_teacher_Qwen2.5-14B_student_Qwen2.5-7B_msmacroo_0.1_2/checkpoint-1400
   \\

   Qwen2.5-14B  & Qwen1.5-1.5B  & 73.63 &	65.85&	83.78	&44.33	&50.81&	34.83&	48.40&	73.30	&28.33	&36.26	&50.00
   %rankLLaMA_teacher_Qwen2.5-14B_student_Qwen2.5-1.5B_msmacroo_0.1_2/checkpoint-1000
   \\
    Qwen2.5-14B  & Qwen1.5-3B  & 73.47&	66.42	&83.42	&45.08	&49.24	&35.59	&49.49&	74.74&	27.27&	36.81&	50.20
   %rankLLaMA_teacher_Qwen2.5-14B_student_Qwen2.5-3B_msmacroo_0.1_2/checkpoint-800
   \\
\midrule
   LLaMA2-13B  & LLaMA-3.2-3B  &  74.49& 	69.02& 	82.91& 	45.28& 	48.99	& 36.17	& 50.93	& 75.48	& 29.14& 	35.78& 	50.58
%rankLLaMA_teacher_13b_student_LLaMA-3.2-3B_msmacroo_0.1_2/checkpoint-11400
\\
LLaMA2-13B  & LLaMA-3.2-1B  & 	72.82&	68.08&	79.00	&43.60	&47.65	&35.79	&47.46&	75.88&	31.21	&37.18	&49.72

%%rankLLaMA_teacher_13b_student_1b_msmacroo_0.1_2/checkpoint-1200	
\\

	LLaMA2-13B  & LLaMA-3.1-8B  & 74.50	&  68.71	&  84.14	& 46.27	& 51.71& 36.57	& 52.43	& 77.39	& 29.91	& 37.23	& 51.95

    %LLaMA8_passage2	rankLLaMA_teacher_13b_student_8b_msmacroo_0.1_2	checkpoint-1000
\\
    \bottomrule
    %%%%%%%%% all these from leonardo server 
\end{tabular}
\caption{nDCG@10 performance (in percentage) of \DeAR-Pointwise with different teacher and student model pairs using binary cross-entropy across TREC and BEIR datasets.}
\label{tab:diffmodels_app}
\end{table*}

\end{document}